\documentclass{bmvc2k}

\title{Deep Motion Blind Video Stabilization}

\def\thefootnote{*}\footnotetext{Equal contribution. $\dagger$Corresponding author.}

\addauthor{Muhammad Kashif Ali\thefootnote{}}{kashifali@hanyang.ac.kr}{1}
\addauthor{Sangjoon Yu\thefootnote{}}{kiddyu1991@gmail.com}{1}
\addauthor{Tae Hyun Kim$\dagger$}{taehyunkim@hanyang.ac.kr}{1}

\addinstitution{
Department of Computer Science\\
Hanyang University\\
Seoul, South Korea}

\runninghead{Ali, Yu, Kim}{Deep Motion Blind Video Stabilization}

\newcommand{\figref}[1]{Figure~\ref{#1}}

\begin{document}

\maketitle

\begin{abstract}
Despite the advances in the field of generative models in computer vision, video stabilization still lacks a pure regressive deep-learning-based formulation. Deep video stabilization is generally formulated with the help of explicit motion estimation modules due to the lack of a dataset containing pairs of videos with similar perspective but different motion. Therefore, the deep learning approaches for this task have difficulties in the pixel-level synthesis of latent stabilized frames, and resort to motion estimation modules for indirect transformations of the unstable frames to stabilized frames, leading to the loss of visual content near the frame boundaries. In this work, we aim to declutter this over-complicated formulation of video stabilization with the help of a novel dataset that contains pairs of training videos with similar perspective but different motion, and verify its effectiveness by successfully learning motion blind full-frame video stabilization through employing strictly conventional generative techniques and further improve the stability through a curriculum-learning inspired adversarial training strategy. Through extensive experimentation, we show the quantitative and qualitative advantages of the proposed approach to the state-of-the-art video stabilization approaches. Moreover, our method achieves $\sim3\times$ speed-up over the currently available fastest video stabilization methods.
\end{abstract}

%-------------------------------------------------------------------------
\section{Introduction}

\begin{figure*}[t]
    \centering
    \includegraphics[width=0.8\textwidth]{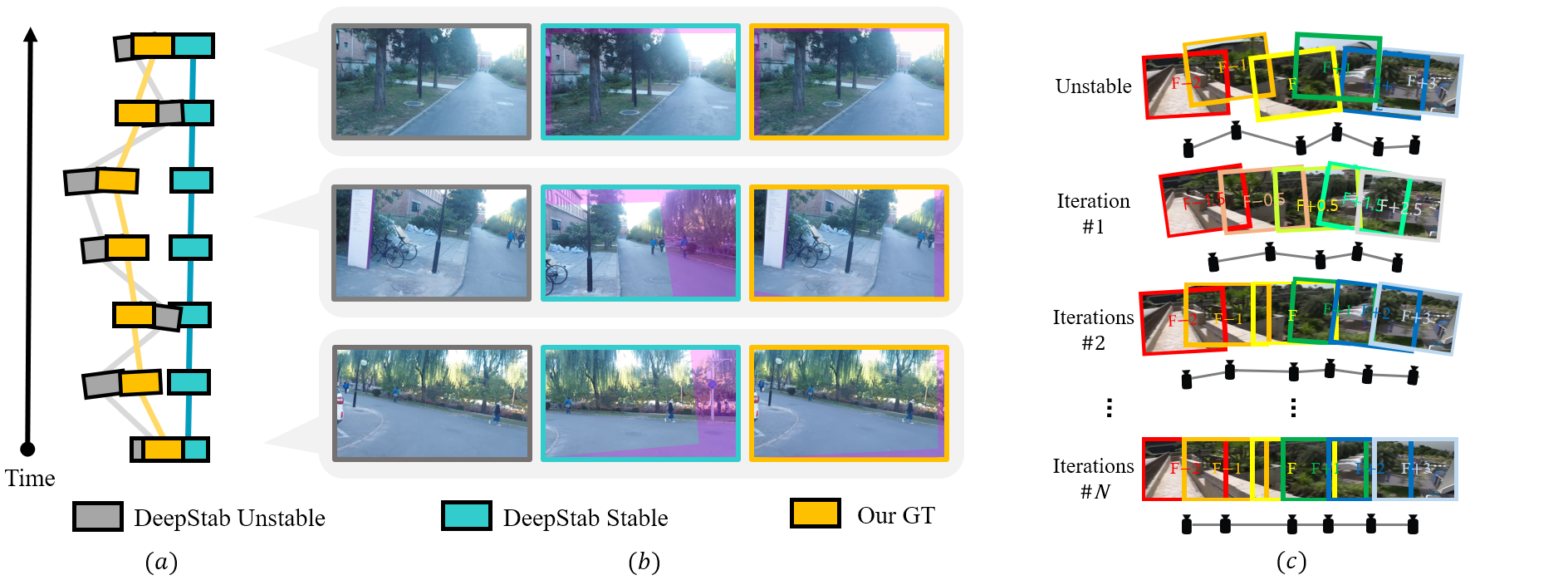}
    \caption{(a) An illustration of the perspective mismatch in the DeepStab dataset and our proposed Dataset Generation Pipeline (DGP). (b) Large non-overlapping regions (transparent purple zone) present in the DeepStab dataset~\cite{wang2018deep} along with the minimized perspective difference in our dataset. (c) A visual description of iterative frame interpolation leading to visual stability and smooth camera trajectory.}
    \label{fig:fig2}
\end{figure*}
The prevalent integration of high-quality cameras in hand-held devices, has enabled the general population to record the memorable moments of their life, but it still requires professional equipment to record stable videos. Thus, considerable literature has been devoted to solving the video stabilization problem. Despite the advances in the generative deep learning models, there is still a long way to go for deep-learning-based approaches to truly take over in video stabilization from the traditional reconstructive  feature-tracking~\cite{liu2014steadyflow,liu2016meshflow} and trajectory optimization~\cite{grundmann2011auto, liu2013bundled} methods. 

Recently, Wang et al.~ \cite{wang2018deep} released the DeepStab dataset, which is the first large-scale dataset for video stabilization.  This dataset is captured with two synchronized cameras placed on a contraption fixed around the base of a mechanical stabilizer. The camera placed on the physical stabilizer captures the stable video while the camera on the contraption rotates freely along the stabilizer and records unstable videos. Due to the rotational motion of the unstabilized camera, both of the recorded videos often contain a significant non-overlapping field-of-view, and a perspective mismatch (as shown in \figref{fig:fig2}). This inconsistency in the perspective makes it difficult for the models to learn the direct pixel-level spatio-temporal relations of unstable videos to their stable counterparts. Thus, video stabilization is generally defined with the help of dense optical flow estimation modules and the networks learn to warp the original frames instead of synthesizing them~\cite{yu2020learning}. This warping generally entails a substantial cropping near the frame boundaries and temporal distortions in the stabilized videos. To overcome this problem, we provide a new dataset by extending and improving the idea of iterative frame interpolation leading to smooth motion trajectories as presented in~\cite{choi2020deep} to generate stable and unstable training videos which virtually share the same perspective (highlighted in yellow in \figref{fig:fig2} (b)). Through our experiments with the proposed dataset, we attempt to declutter and relieve the dependence on motion-awareness in the formulation of video stabilization pipelines, and demonstrate that full-frame video stabilization can be formulated with conventional network architectures and modules without explicit motion awareness.
In addition, we further propose a contrastive motion loss and a temporal adversarial training strategy to produce more stable and temporally consistent full-frame videos. Our proposed stabilization network compares favorably to the currently available motion-aware solutions, and we summarize our contributions as follows:\\

\begin{itemize}
    \item \textbf{Unsupervised dataset generation:} we introduce an unsupervised and extensible video-frame-interpolation-based strategy to produce equi-perspective stabilized videos from unstable videos captured from hand-held devices.
    \item \textbf{Motion blind full-frame video stabilization:} we declutter the overly complex video stabilization formulation and propose the first ever motion blind deep stabilization network with the help of the proposed equi-perspective dataset.
    \item \textbf{Curriculum Learning strategy:} we present a targeted sequential learning strategy where we allow the same network to focus on multiple aspects of stabilization in different stages.
\end{itemize}

%$\bullet$ {Unsupervised dataset generation:} we introduce an unsupervised and extensible video-frame-interpolation-based strategy to produce equi-perspective stabilized videos from unstable videos captured from hand-held devices.\\
%$\bullet$ {Motion blind full-frame video stabilization:} we declutter the overly complex video stabilization formulation and propose the first ever motion blind deep stabilization network with the help of the proposed equi-perspective dataset.

%$\bullet$ {Curriculum Learning strategy:} we present a targeted sequential learning strategy where we allow the same network to focus on multiple aspects of stabilization in different stages.

\section{Related Work}
Liu et al.~\cite{liu2009content} proposed a 3D approach for this task, in which camera poses along with feature tracks were reconstructed in the 3D space, and the feature positions were projected along smoothened camera poses, whereas, Smith et al.~\cite{smith2009light} employed depth-aware cameras to do the same.
However, these global 3D approaches cannot properly handle dynamic scenes including moving objects, and thus 2D transformations (e.g., homography) become more popular in video stabilization methodologies.
In general, these 2D methods rely on tracking prominent features and stabilizing their trajectories along the motion path. The results produced by these methods generally need cropping around the borders and up-scaling to retain the original resolution of the input video. In addition, Buehler et al.~\cite{buehler2001non} estimated the camera positions through shaky videos and rendered the frames at smoothened camera positions using a non-metric Image Based-Rendering method. Matsushita et al.~\cite{matsushita2006full} and Gleicher et al.~\cite{gleicher2008re} used simplistic 2D transformation mechanisms to warp the original frames. Whereas, Liu et al.~\cite{liu2013bundled} introduced a grid-based warping of frames for smoothing the feature trajectories. Grundmann et al.~\cite{grundmann2011auto} proposed an L1-based cost functions for obtaining optimal camera trajectory for stabilized feature tracks, whereas, Liu et al.~\cite{liu2011subspace} proposed a similar approach but employed the eigen-trajectory smoothing technique. Wang et al.~\cite{wang2013spatially} and Goldstein et al.~\cite{goldstein2012video} also approached this task with optimization-based models to acquire feature tracks and camera position using epipolar geometry.

All of these methods relied heavily on feature tracks and ignored the underlying relation of independent motion of multiple objects in consecutive frames, which compelled Liu et al.~\cite{liu2016meshflow, liu2014steadyflow} to investigate applications of optical flow in the field of video stabilization. Their studies helped understanding the importance of inter-frame motion estimation in video stabilization and paved the path for modern video stabilization methods. Thus, Yu and Ramamoorthi et al.~\cite{yu2019robust, yu2020learning} and Choi et al.~\cite{choi2020deep} employed dense optical flow estimation modules to warp the neighboring frames to obtain smoother and better-quality videos. In particular, Yu and Ramamoorthi et al.~\cite{yu2019robust} proposed a scene-specific optimization approach that estimates dense motion to optimize the network weights for each video and extended their approach in~\cite{yu2020learning} to a generalized framework capable of handling complex situations including (de)occlusion and non-linear motion through warp fields.

Two pioneering methods implicitly using motion flow were proposed in~\cite{xu2018deep,wang2018deep}. These methods employ generative adversarial networks and spatial transformer networks to learn the inter-frame motion and warp the frames for video stabilization. Wang et al.~\cite{wang2018deep} proposed the DeepStab dataset and attempted to find a possible solution for video stabilization with a Siamese network containing a pre-trained ResNet50 model.
Another attempt to train a pure image-based stabilizer using the DeepStab dataset without motion estimation was discussed in~\cite{yu2020learning} and was termed as an “essential over-fitting task", because using this dataset for training can lead to the network learning an entirely different perspective of the same scene without the presence of any correlation or information about the perspective in the input unstable video frames.

Meanwhile, a new stabilization network based on dense optical flow estimation and video frame interpolation was proposed in~\cite{choi2020deep} called DIFRINT. They achieve temporal stability by rendering interpolated frames between unstable frames and obtain full-frame video stabilization results. %We observe that their optical flow estimator fails when there is complex inter-frame motion as shown in \figref{fig:fig3}.
These attempts have helped us in pinpointing the shortcomings of the DeepStab dataset and have encouraged us to propose an equi-perspective dataset which can simplify the task of video stabilization.%understand the importance of an often neglected factor (perspective) in the dataset for the video stabilization task and encouraged us to formulate video stabilization as a supervised learning task independent of motion estimation.

\section{Dataset Generation Pipeline (DGP)} \label{data_gen}
\begin{figure}[t]
    \includegraphics[width=0.45\textwidth]{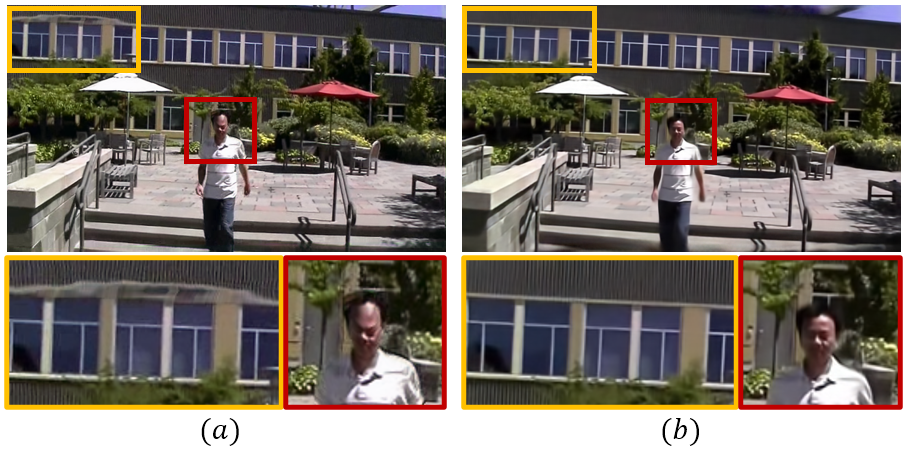}
    \centering
    \caption{(a) Interpolated frame by DIFRINT~\cite{choi2020deep}. (b) Interpolated frame by CAIN~\cite{choi2020channel}.}
    \label{fig:fig3}
\end{figure}

\begin{figure}[b]
    \centering
    \includegraphics[width=0.5\textwidth]{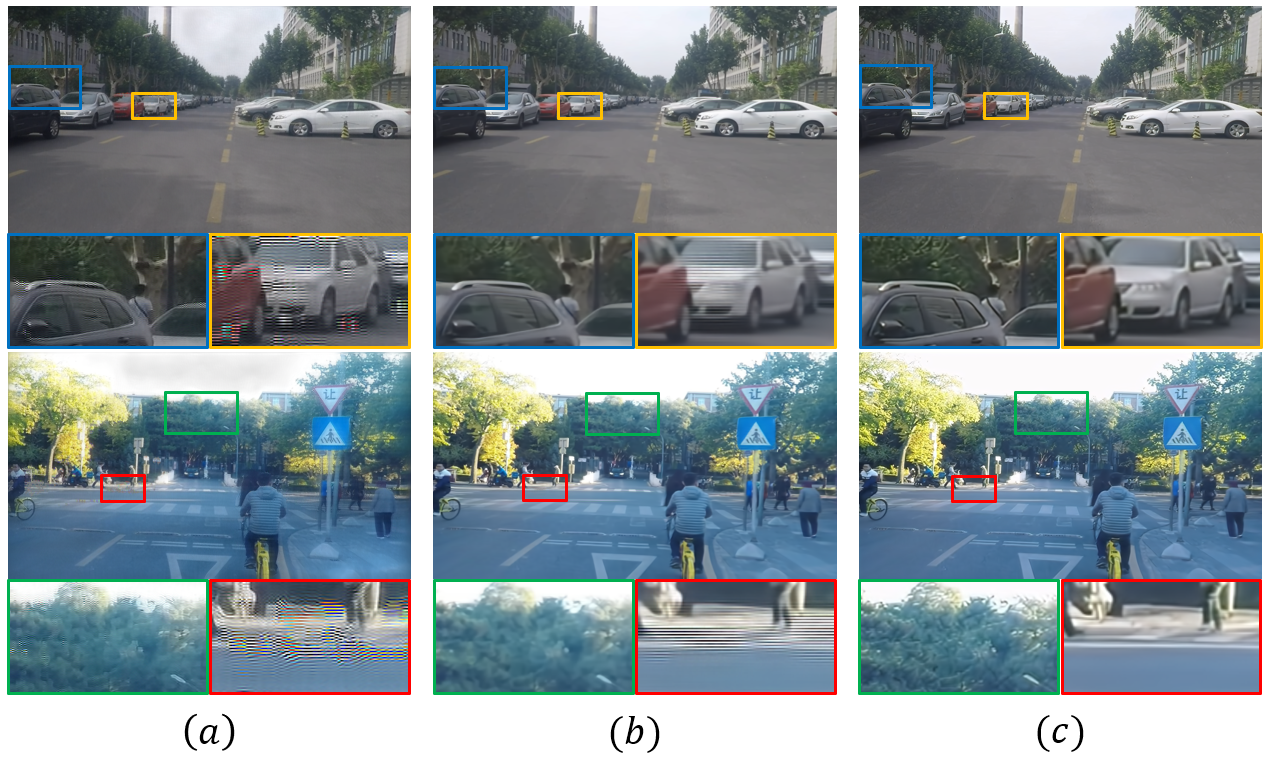}
    \caption{Comparison of frame interpolation methods used in iterative arrangement (20 iterations). (a) SepConv~\cite{niklaus2017video}. (b) CAIN~\cite{choi2020channel}. (c) Proposed DGP.}
    \label{fig:fig4}
\end{figure}
The generation of a labeled dataset for video stabilization is a challenging task. Before finalizing our dataset generation pipeline, we experimented with various techniques to pinpoint the missing link that hinders motion blind formulation of this task. Our experiments included training the same network on DeepStab~\cite{wang2018deep} dataset and a dataset generated through random affine transforms. Through this experiment, we observed that the network trained with the synthetic dataset learns to better stabilize videos than the network trained on the DeepStab~\cite{wang2018deep} dataset. In order to minimize the effect of large non-overlapping regions, we also experimented with downscaling the DeepStab~\cite{wang2018deep} to an eighth of its original frame size. This downscaling operation reduced the overall inter frame motion. Even with these downscaled frames, we were unable to learn meaningful stabilization. Through these experiments, we concluded that learning the high-level reasoning for stabilization does not just require a minimized non-overlapping region but it also requires the target and input videos to share similar perspective in order to properly find correspondences. This is generally avoided in the motion aware techniques with the help of dense optical flow estimation which helps the model to differentiate between local and global motion present between the frames along with an abstract sense of jerkiness in unstable videos. In order to generate a new large-scale video stabilization dataset that fulfills these requirements, we draw motivation from DIFRINT~\cite{choi2020deep}, which is a frame-interpolation-based stabilization network. Specifically, the first part of their network is similar to the conventional video frame interpolation networks. Their intuition for incorporating this network in their pipeline was to achieve temporal stability by reconstructing frames with high-frequency jerks. Their pseudo frame interpolation network is trained for video stabilization, with a synthetic dataset generated using random affine transformations. 
This synthetic dataset lacks the complexity of real motion in dynamic scenes.
Thus, this network faces difficulties in handling the differently moving objects in real-world scenarios, and often results in undesirable wobble like artifacts as shown in \figref{fig:fig3} (a). Notably, in the modern frame interpolation methods~\cite{choi2020channel, niklaus2017video}, this is compensated by training the networks on real-world videos containing complex inter-frame local and global motions as shown in \figref{fig:fig3} (b). Based on these observations, we deduce that conventional video frame interpolation models (specifically trained for this task) can outperform DIFRINT~\cite{choi2020deep} in handling dynamic motion scenarios and produce stabler and higher quality videos.

In general, video frame interpolation methods behave like a low-pass filter and produce a middle frame by blending the neighboring frames. This generation of the middle frame can remove high-frequency jitter present between the alternate frames. In an iterative arrangement, this approach can take into account the relative motion of all the frames present in the sequence, and generate a temporally consistent sequence free from sudden high-frequency jerks. It is worth noting that DIFRINT~\cite{choi2020deep} enforces stability by skipping intermediary frames in their stabilization pipeline and generates the skipped frames with an assumption that the intermediary frames lie along a straight line and the inter-frame motion is strictly linear. These assumptions enforce temporal stability in fairly lesser iterations, but result in undesirable artifacts around the depth boundaries of distinctly moving objects, and a jagged progression in the stabilized videos. First, to solve these problems we propose a new dataset generation technique, which instead of skipping frames, generates the intermediate frames and uses these intermediate frames to reconstruct the original sequence which preserves the original progression of the video sequence. Secondly, we propose a refinement network which restores the integrity of iteratively generated data.

$\bullet$ \textbf{Iterative Frame Interpolation:} For the first part of our dataset generation pipeline, we tested two state-of-the-art frame interpolation methods SepConv~\cite{niklaus2017video} and CAIN~\cite{choi2020channel} in an iterative arrangement and opted to use CAIN~\cite{choi2020channel} as it produced better quality frames in our experiments (presented in \figref{fig:fig4}). Unlike DIFRINT~\cite{choi2020deep}, we use our frame interpolator without any rigid assumptions about the nature of inter-frame motion, and only compensate for high-frequency jerks through our iterative frame interpolation pipeline. This iterative stability comes at the cost of visual distortions and artifacts. Although the generated frames include lesser high-frequency jerks, various artifacts such as blur and color distortions are generated as shown in \figref{fig:fig4}. To overcome these problems, we additionally introduce a refinement network that restores the visual integrity of the generated frames.

\begin{figure*}[b]
    \centering
    \includegraphics[width=1.0\textwidth]{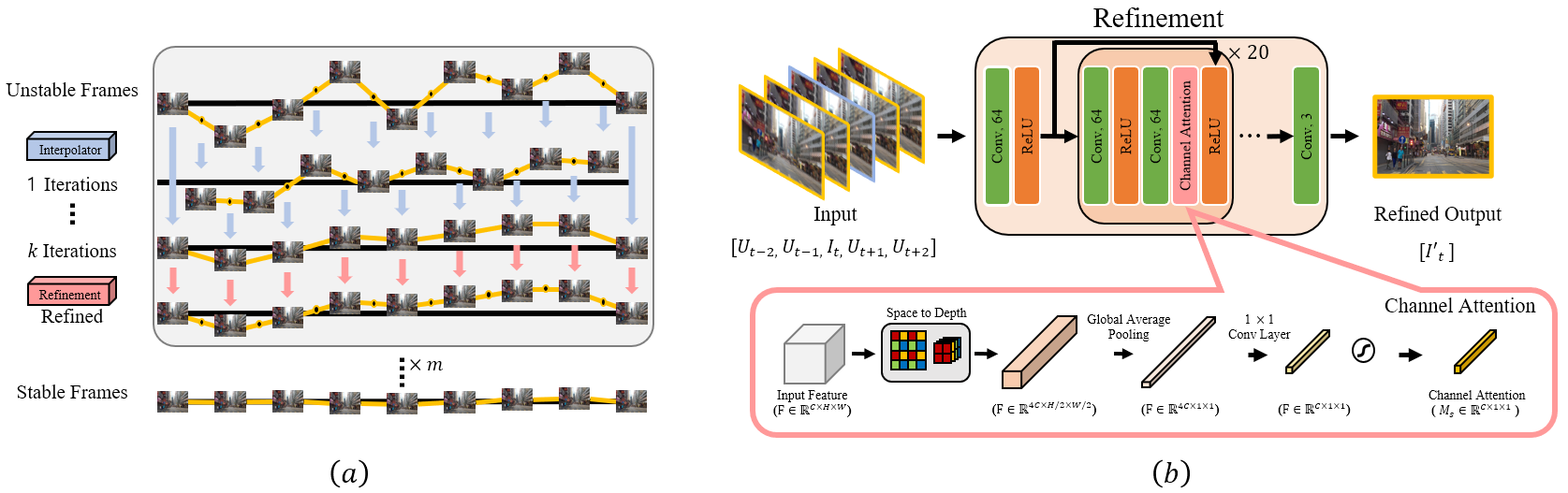}
    \caption{(a) Proposed DGP for video stabilization with integrated the refinement network. (b) The architecture and inference strategy of the refinement network.}%, it takes in original unstable but sharp frames $(U_{t-2},U_{t-1},U_{t+1},U_{t+2})$ and one interpolated frame $I_{t}$ to render a clean frame.}
    \label{fig:fig5}
\end{figure*}

$\bullet$ \textbf{Refinement Network:}
To remove the artifacts introduced by the iterative frame interpolations, we introduce a \emph{refinement} network. 
%This network merges and takes advantage of the visual quality improvement techniques and modules proposed in~\cite{choi2020channel, lim2017enhanced, ma2020structure, nah2017deep, sajjadi2017enhancenet} for their proven success in generating visually pleasing images.
Our refinement network is based on ResNet~\cite{he2016deep} with a modified version of the channel attention module from CAIN~\cite{choi2020channel} (as shown in \figref{fig:fig5}).
Our modified attention module treats the features with a succession of space-to-depth operations followed by global average pooling, and a $\emph{1} \times \emph{1}$ convolution layer. The output of this layer is passed through a sigmoid function and then multiplied (element-wise) to the input features. 
We observed that the original frames contain unaltered high-quality regions necessary to restore the degraded interpolated frames. Thus, this network takes in an interpolated frame with its neighboring original (unstable) frames as input, and generates the restored version of the interpolated frame. Through customized losses and training strategy, we ensure that the network does not alter the spatial relations of the content present in the interpolated frame and only targets the artifacts introduced by the iterative interpolation. A comparison of the refined results with the interpolated results is provided in \figref{fig:fig4}. Please refer to our supplementary material for the detailed dataset generation, formulation, configuration and training and testing strategies of this network.
For the finalized DGP, the refinement network was integrated within the iterative frame interpolation pipeline. We introduced a refinement step after every \emph{k} iterations of the frame interpolation network (i.e., CAIN), and repeated this setting for a definite number of times \emph{m} (as illustrated in \figref{fig:fig5}) to acquire temporally stable and high-quality frames as shown in~\figref{fig:fig4} (c). In our experiments, we select \emph{k} and \emph{m} to be 4 and 5, respectively. Due to the space limitation, we present a visual ablation study to justify our choices of parameters \emph{k} and \emph{m} in the accompanied supplementary material. We utilize unstable videos from the DeepStab dataset along with videos acquired through the internet to generate our final video stabilization dataset.

\section{Learning Motion Blind Video Stabilization} \label{methodology}
%\vspace{-2Ex}
\begin{figure*}[t]
    \centering
    \includegraphics[width=0.8\textwidth]{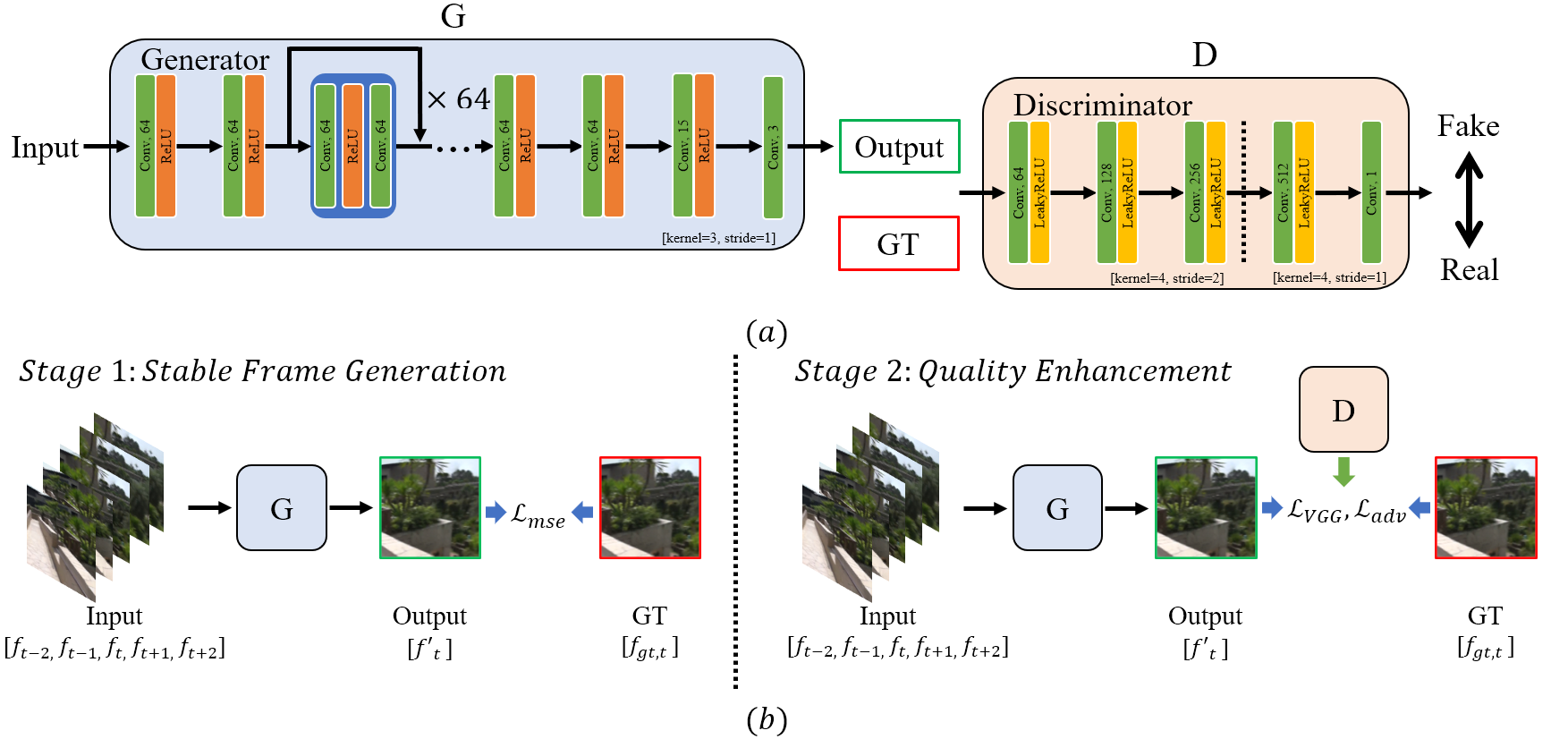}
    \caption{(a) Network architecture of the proposed stabilization network along with the discriminator for training. (b) Training strategy for Stage 1 and Stage 2 (left to right)}
    \label{fig:fig7}
\end{figure*}
%\vspace{-3Ex}
\subsection{Re-formulation of Video Stabilization}
The conventional deep video stabilization methodologies formulate the task of video stabilization with explicit motion estimation modules. Generally, these modules further complicate an already convoluted formulation with additional steps to process the calculated motion flows as described in~\cite{yu2020learning}. Contrary to the normal convention, we propose a simplistic and straightforward formulation of the video stabilization through our generated dataset. The minimized perspective mismatch in our dataset assists the model to focus on the spatial relations between the stable and unstable videos.
Various model architectures like U-Net~\cite{ronneberger2015u} and ResNet~\cite{he2016deep} structures were tested before finalizing the baseline architecture for this task. We employ a modified version of a super-resolution network, ENet~\cite{sajjadi2017enhancenet} (based on ResNet architecture) for our stabilization network. A very deep ResNet based architecture taking multiple input frames allows the network to exploit spatio-temporal information along with an extended receptive field as described in~\cite{wang2018deep}. The architecture of the proposed network for video stabilization is shown in~\figref{fig:fig7}. Our model takes in five consecutive unstable frames and produces the stabilized version of the middle frame. The number of input frames used for the stabilization network was evaluated empirically. %The upcoming sections will describe the training strategy of the proposed stabilization network.

\subsection{Training Strategy}
Inspired from the ideas presented in~\cite{elman1993learning, xu2018deep}, we divide the task of learning motion blind video stabilization in to its different components and allow the model to focus on only one task at a time. With this strategy we can learn all the aspects of video stabilization through the same network. In the first two stages of our training, namely, \emph{Stable Frame Generation} and the \emph{Quality Enhancement}, we purposefully employ conventional methodologies to verify the effectiveness of the proposed dataset and the importance of equi-perspective training samples for this task. Whereas, in the third stage, \emph{Strengthening}, we let the model focus on learning the abstract reasoning for improving the stability and temporal consistency.

\subsubsection{Stage 1: Stable Frame Generation} \label{Stage1}
In this stage, we train the network with a specific goal of generating stable frame \emph{$f^{\prime}_{t}$} from the five unstable input frames $(f_{t-2}, f_{t-1}, f_t, f_{t+1,}, f_{t+2})$ (as presented on the left side of \figref{fig:fig7} (b)). During this stage, the perceptual quality of the generated frames is purposefully ignored as it can be enhanced in the upcoming stages with the help of an adversarial training strategy and a perceptual loss. During our experiments, it was observed that introducing a quality improvement loss at this stage significantly increased the convergence time. Therefore, at this stage, the model is trained with a specific goal of learning only the high-level reasoning necessary to justify the generated output frame $f^{\prime}_{t}$ from the input unstable frames. 
In this stage, we train the stabilization network with the $\mathcal{L\textsubscript{2}}$-based reconstruction loss as,
\begin{equation}
\mathcal{L} = {\left \| {f}'_{t} - {f}_{gt, t} \right \|}_{2}^{2},
\label{eq_l2}
\end{equation}
where \emph{f\textsubscript{gt,t}} is the frame acquired through our proposed DGP (described in Sec. \ref{data_gen}).

\subsubsection{Stage 2: Quality Enhancement} \label{Stage2}
After the convergence with the $\mathcal{L\textsubscript{2}}$ reconstruction loss in~(\ref{eq_l2}), the results produced by the network are stable but quite blurry \footnote{The quality of the results produced at this stage can be assessed through the supplementary text}. The perceptual quality of these stable but blurry frames can be improved by fine-tuning the network with a perceptual and an adversarial loss for their proven success in enhancing the visual quality of degraded images~\cite{kupyn2017deblur,kupyn2019deblur,nah2017deep}. 
The primary loss used during this stage is a VGG based loss defined as follows:
\begin{equation}
    %\mathcal{L} = {\left || \phi ( {f}'_{t} ) -\phi \left ( {f}_{gt, t} \right ) \right ||}_{2}^{2} ,
    \mathcal{L\textsubscript{content}} = {\left \| \phi ( {f}'_{t} ) -\phi ( {f}_{gt, t}  ) \right \|}_{2}^{2} ,
\end{equation}
%where \emph{$\phi\left ( {f}'_{t} \right )$} are the VGG features extracted from the generated frame \emph{$f^{\prime}_{t}$}, and \emph{$\phi \left ( {f}_{gt, t} \right )$} are those extracted from ground-truth target frame \emph{f\textsubscript{gt,t}}. For the purpose of feature extraction, a pre-trained VGG-19 is used up to the \emph{relu$\textunderscore$3$\textunderscore$3} layer. This loss ensures the preservation of high-level visual cues present as proposed in~\cite{johnson2016perceptual}.
where \emph{$\phi\left (\cdot \right )$} represents the \emph{relu$\textunderscore$3$\textunderscore$3} layer of a pre-trained VGG-19 network. This loss ensures the preservation of high-level visual cues present as proposed in~\cite{johnson2016perceptual}. In addition to the perceptual loss, we also employ an adversarial training schema in this stage. The discriminator used in our work is shown in ~\figref{fig:fig7} (a). It is a simple feed-forward network inspired by the discriminator used in~\cite{kupyn2018deblurgan} with alternating convolution and Leaky-Relu operations. The final loss for training in this stage is given by the following equation,
\begin{equation}
    \mathcal{L} = \mathcal{L\textsubscript{content}} + \lambda \cdot \mathcal{L\textsubscript{adv}} ,
\end{equation}
Here, $\mathcal{L}_{adv}$ is the adversarial loss and $\lambda$ denotes a user-parameter that controls the contribution of the adversarial loss in the optimization step. As for the adversarial loss we utilize WGAN-GP~\cite{NIPS2017_892c3b1c} for its success in similar quality improvement tasks such as~\cite{kupyn2018deblurgan, kupyn2019deblur}.
%We provide a visual ablation comparison to highlight the results produced before-and-after \emph{Stage 2} in~\figref{fig:fig8} (c).
%It is evident from this comparison that the results produced by the second stage contain sharp edges while preserving the learned spatial relations during \emph{Stage 1}. 
A brief inter-stage ablation study is provided in the accompanied supplementary material.
At this stage we verify and prove the effectiveness of the proposed dataset and show that, the video stabilization pipelines can be simplified with the help of our proposed dataset containing pairs of stable and unstable training videos with a minimized perspective difference.

\subsubsection{Stage 3: Strengthening} \label{Stage3}
Wobble effect (as highlighted in \figref{fig:fig3} (a)) is quite common in digitally stabilized videos. This effect occurs due to the motion compensation, and it can be minimized in motion aware approaches at the cost of stability. Since our model does not contain any explicit motion estimation module, we address this issue with the help of specialized losses for this task. During our experiments we observed that the natural video sequences do not contain these artifacts. Therefore, we propose a temporal discriminator that can differentiate between a natural sequence and an artificially generated one. With this intuition, we introduce a secondary discriminator which takes in 16 sequential frames of the generated videos along with the corresponding DeepStab~\cite{wang2018deep} stable frames, and encourages the proposed stabilization network to generate wobble free frames. We also employ a contextual~\cite{mechrez2018contextual} and a perceptual loss~\cite{johnson2016perceptual} between the generated and the unstable frames for content preservation. 
In addition to these losses, we also propose a contrastive motion loss to enhance stability. This loss uses an off the shelf pre-trained Video ResNet-18 for action recognition as proposed in~\cite{hara2017learning} to produce embeddings for the generated video sequences along with the corresponding DeepStab stable and unstable sequences. These embeddings are then used with a triplet loss~\cite{dong2018triplet}. The embeddings for the DeepStab stable, unstable and our generated sequences are used as anchor, negative and positive embeddings respectively. This loss minimizes the distance between the positive and anchor while maximizing the distance between the anchor and the negative embeddings. During the experimentation, an increase of 2-3\% in the stability values from the Stage 2 network was observed by the introduction of this loss. The final loss for training at this stage is given by the following equation,
\begin{equation}
    \mathcal{L} = {\lambda}_{1} \cdot \mathcal{L\textsubscript{$\phi$}} + {\lambda}_{2} \cdot \mathcal{L\textsubscript{CX}} + {\lambda}_{3} \cdot \mathcal{L\textsubscript{td}} + {\lambda}_{4} \cdot \mathcal{L\textsubscript{id}} + {\lambda}_{4} \cdot \mathcal{L\textsubscript{cml}},
\end{equation}

Here, $\mathcal{L}_{\phi}$, $\mathcal{L}_{CX}$, $\mathcal{L}_{td}$, $\mathcal{L}_{id}$ and $\mathcal{L}_{cml}$ represent, perceptual, contextual, temporal discriminator, image discriminator and contrastive motion loss, respectively. Here, ${\lambda}_{n}$ represent the controlling hyperparameters. Due to the space limitation, the details of the above-mentioned losses and the implementation are provided in the supplementary material. 

%It is noteworthy that during our experimentation phase, we experimented with training the stabilization network with all the finalized losses at once, but failed to acquire similar quantitative and quantitative results.

\section{Results}
\subsection{Quantitative Results}
\begin{figure}[h]
    \centering
    \includegraphics[width=0.9\textwidth]{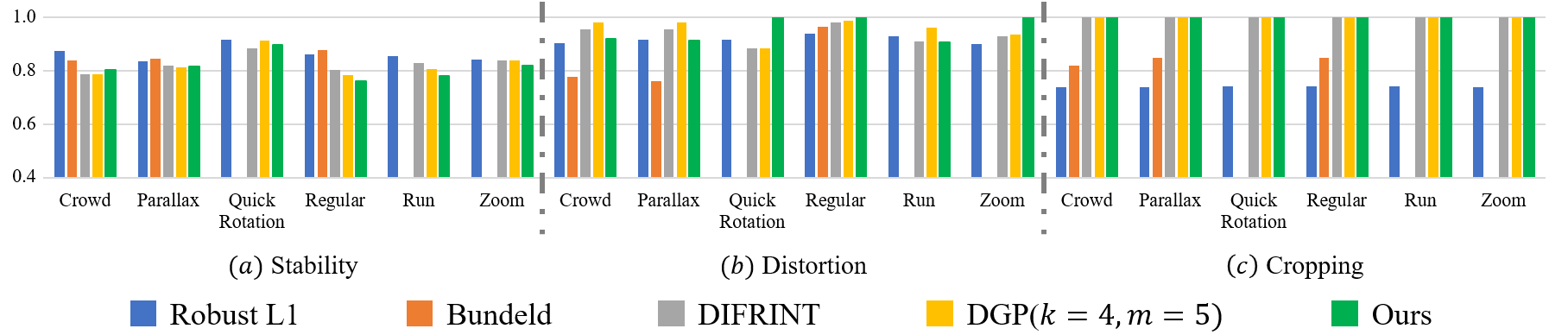}
    \caption{Quantitative comparison of Robust L1~\cite{grundmann2011auto}, Bundled~\cite{liu2013bundled}, DIFRINT~\cite{choi2020deep}, DGP (\emph{k=4, m=5}) and our stabilization network.}
    \label{fig:quan_result}
\end{figure}
%\vspace{-2Ex}
We evaluate the performance of the proposed method quantitatively in terms of stability, distortion and cropping metrics as suggested by~\cite{liu2013bundled} on the 6 categories of videos presented in the NUS dataset~\cite{liu2013bundled}. The provided results (\figref{fig:quan_result}) are averaged over each category.\\
%\begin{itemize}
%\item \textbf{Stability:} This metric defines the stability in terms of frequency component analysis. To calculate this metric, the feature trajectories are analyzed in the frequency domain as described in~\cite{liu2013bundled}. It is worth emphasizing that this metric does not take into account the quality of input videos and blurry results are also perceived stable through this metric.
%\item \textbf{Distortion:} This metric evaluates the anisotropic homography of the generated frames to the actual unstable frames. The lowest ratio is selected as the final distortion score. A higher score in this metric signifies better preservation of the content.
%\item \textbf{Cropping:} This metric measures the retention of visual information in generated frames through homography calculation between the  generated and the actual frames. A higher score signifies better preservation of the visual information.
%\end{itemize}
$\bullet$ \textbf{Stability:} This metric defines the stability in terms of frequency component analysis. To calculate this metric, the feature trajectories are analyzed in the frequency domain as described in~\cite{liu2013bundled}. It is worth emphasizing that this metric does not take into account the quality of input videos and blurry results are also perceived stable through this metric.\\
$\bullet$ \textbf{Distortion:} This metric evaluates the anisotropic homography of the generated frames to the actual unstable frames. The lowest ratio is selected as the final distortion score. A higher score in this metric signifies better preservation of the content.\\
$\bullet$ \textbf{Cropping:} This metric measures the retention of visual information in generated frames through homography calculation between the  generated and the actual frames. A higher score signifies better preservation of the visual information.

Through \figref{fig:quan_result}, it is evident that the proposed network outperforms the SOTA methods in terms of distortion and cropping and performs competitively in terms of stability on the videos from Crowd, Parallax and Quick Rotation and lags behind in the remaining three categories. This is due to the fact that a large portion of the generated dataset consists of the unstable videos from the DeepStab~\cite{wang2018deep}, which contain the motion profiles similar to the above mentioned three categories. This bias in the results can be minimized by fine-tuning the network on videos containing motion profiles similar to the videos from the remaining three categories. We do not include the quantitative results produced by iterated CAIN~\cite{choi2020channel} as the results contain inconsistent inter-frame artifacts that hinder the calculation of stability score by introducing a new local motion profile in the resulting videos. It is worth noting that the videos generated by the DGP and the iterated CAIN~\cite{choi2020channel} share the same global motion profiles hence the actual stability score for both the methods should be similar. Please note that the \figref{fig:quan_result} does not include the results from~\cite{liu2013bundled} for Quick Rotation, Run and Zoom as it fails to stabilize most of the videos from these categories because of the extremely large non-overlapping regions.

\subsection{Qualitative Results}
\begin{figure*}[h]
    \centering
    \includegraphics[width=0.9\textwidth]{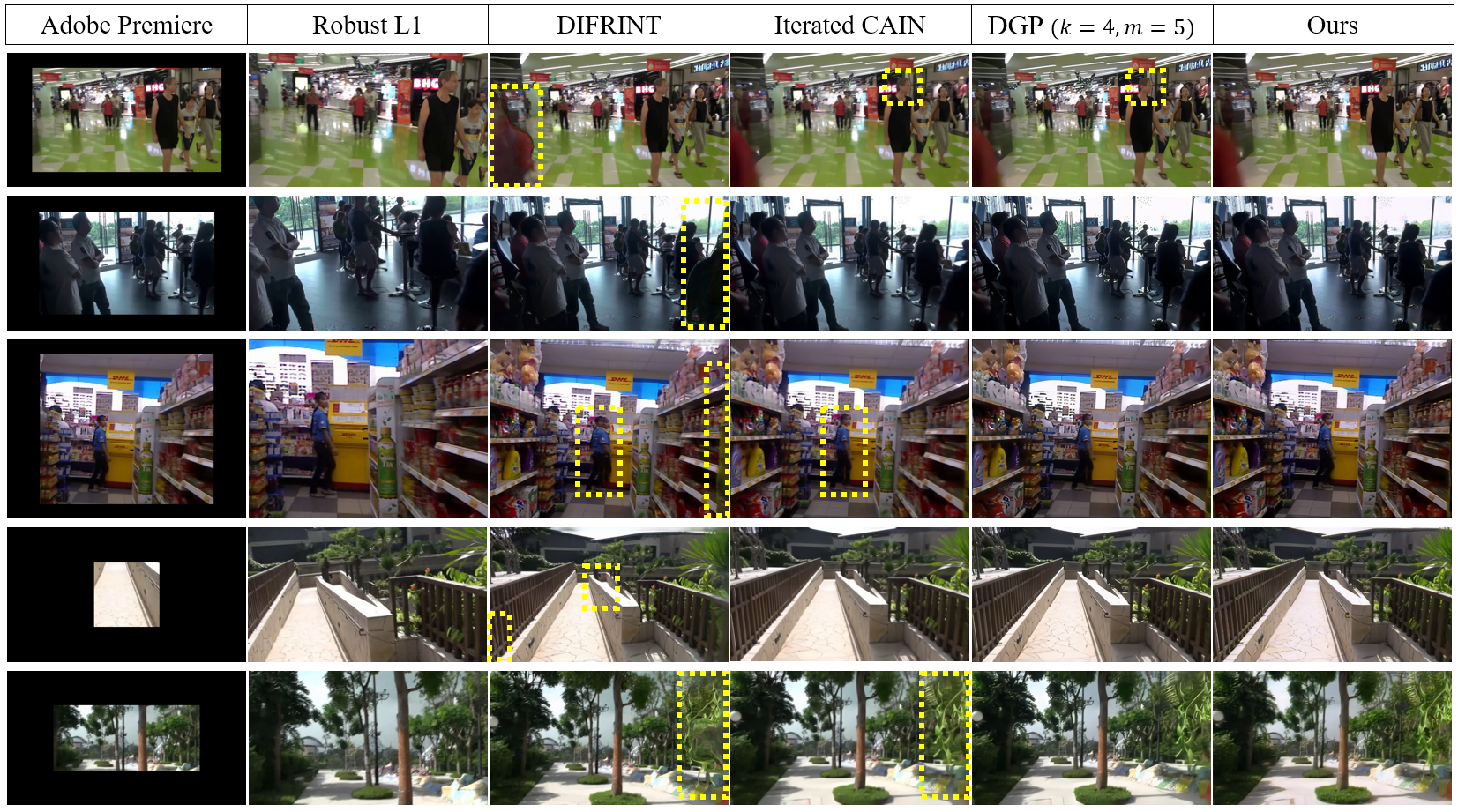}
    \caption{Visual quality comparison of Adobe Premiere 2018 CC, Robust L1~\cite{grundmann2011auto}, DIFRINT~\cite{choi2020deep}, Iterated CAIN~\cite{choi2020channel}, DGP (\emph{k=4, m=5}), and our stabilization network.}
    \label{fig:Quality}
\end{figure*}

For visual quality comparison, we present the results generated by Adobe Premiere 2018 CC, Robust L1~\cite{grundmann2011auto}, DIFRINT~\cite{choi2020deep},  Iterated CAIN~\cite{choi2020channel} (20 iterations), frames generated through our DGP (\emph{k=4, m=5}) and the output from the proposed stabilization network in~\figref{fig:Quality}.
The loss of visual resolution can be clearly seen in the results by Adobe premiere 2018 CC and Robust L1~\cite{grundmann2011auto}. The bounded yellow regions in DIFRINT~\cite{choi2020deep} and iterated CAIN~\cite{choi2020channel} highlight the artifacts caused by both methods. It can be seen from these results that our models (DGP and stabilization network) produce better quality results and preserve the scale and content. The user study and more results are presented in the supplementary material.
%\vspace{-3Ex}
\section{Conclusion} 
%In this paper, we present the first ever motion-blind deep video stabilization solution along with a large-scale dataset for video stabilization. The proposed dataset consists of unstable videos taken from unconstrained environments and stabilized through iterative frame interpolation. Through extensive experiments, we have demonstrated that the proposed stabilization network trained with the proposed dataset compares favorably in terms of stability to the state-of-the-art methods which employ explicit motion estimation modules, and performs exceptionally well in preserving the visual information as well as the resolution.
%\vspace{-1Ex}
In this work, we firstly pinpoint the obstacles that hinder a motion blind video stabilization formulation, and then present the first ever pixel-level synthesis solution for it. To do so, we firstly propose a dataset generation scheme that produces equi-perspective high-quality stable videos through iterative frame interpolation and refinement. Through the generated dataset, and a carefully designed training strategy, we demonstrate that the proposed motion blind video stabilization network compares favorably to the state-of-the-art video stabilization solutions that utilize explicit motion estimation modules, and our proposed model also preserves the visual information as well as the resolution which the currently available methods struggle with.

\section*{Acknowledgement}
This work was supported by Samsung Research Funding Center of Samsung Electronics under Project Number SRFCIT1901-06.

\clearpage
\bibliography{bmvc_review}
\end{document}